\documentclass[12pt,twoside]{article}
\usepackage{latexsym,amsmath,amssymb,hyperref}
\newdimen\paravsp  \paravsp=1.3ex
\def\,{\mskip 3mu} \def\>{\mskip 4mu plus 2mu minus 4mu} \def\;{\mskip 5mu plus 5mu} \def\!{\mskip-3mu}
\def\dispmuskip{\thinmuskip= 3mu plus 0mu minus 2mu \medmuskip=  4mu plus 2mu minus 2mu \thickmuskip=5mu plus 5mu minus 2mu}
\def\textmuskip{\thinmuskip= 0mu                    \medmuskip=  1mu plus 1mu minus 1mu \thickmuskip=2mu plus 3mu minus 1mu}
\textmuskip
\def\beq{\dispmuskip\begin{equation}}    \def\eeq{\end{equation}\textmuskip}
\def\beqn{\dispmuskip\begin{displaymath}}\def\eeqn{\end{displaymath}\textmuskip}
\def\bqa{\dispmuskip\begin{eqnarray}}    \def\eqa{\end{eqnarray}\textmuskip}
\def\bqan{\dispmuskip\begin{eqnarray*}}  \def\eqan{\end{eqnarray*}\textmuskip}
\newenvironment{keywords}{\centerline{\bf\small
Keywords}\begin{quote}\small}{\par\end{quote}\vskip 1ex}
\def\paradot#1{\vspace{\paravsp plus 0.5\paravsp minus 0.5\paravsp}\noindent{\bf\boldmath{#1.}}}
\def\paranodot#1{\vspace{\paravsp plus 0.5\paravsp minus 0.5\paravsp}\noindent{\bf\boldmath{#1}}}
\def\req#1{(\ref{#1})}
\def\eps{\varepsilon}
\def\epstr{\epsilon}                    
\def\nq{\hspace{-1em}}

\def\frs#1#2{{^{#1}\!/\!_{#2}}}
\def\SetR{I\!\!R}
\def\SetN{I\!\!N}
\def\qmbox#1{{\quad\mbox{#1}\quad}}
\def\E{{\bf E}}                         
\def\P{{\rm P}}                         

\def\t{\theta}
\def\l{{\ell}}                          
\def\M{{\cal M}}
\def\X{{\cal X}}
\def\Y{{\cal Y}}                        
\def\o{\omega}
\def\Km{K\!m}
\def\Loss{\mbox{Loss}}
\def\u#1{{\it #1}}

\begin{document}

\title{\vskip 2mm\bf\Large\hrule height5pt \vskip 4mm
Universal Learning Theory
\vskip 4mm \hrule height2pt}
\author{{\bf Marcus Hutter}\\[3mm]
\normalsize RSISE$\,$@$\,$ANU and SML$\,$@$\,$NICTA \\
\normalsize Canberra, ACT, 0200, Australia \\
\normalsize \texttt{marcus@hutter1.net \ \  www.hutter1.net}
}
\date{February 2011}
\maketitle

\begin{abstract}
This encyclopedic article gives a mini-introduction into the
theory of universal learning, founded by Ray Solomonoff in the
1960s and significantly developed and extended in the last
decade. It explains the spirit of universal learning, but
necessarily glosses over technical subtleties.
\vspace{2ex}\def\contentsname{\centering\normalsize Contents\\}
{\parskip=-2.7ex\tableofcontents}
\end{abstract}


\newpage
\section{Definition, Motivation and Background}\label{secIntro}

Universal (machine) learning is concerned with the development and
study of algorithms that are able to learn from data in a very large
range of environments with as few assumptions as possible. The
class of environments typically considered includes all computable
stochastic processes. The investigated learning tasks range from
\u{inductive inference}, sequence prediction, sequential decisions, to
(re)active problems like \u{reinforcement learning}
\cite{Hutter:04uaibook}, but also include \u{clustering}, \u{regression},
and others \cite{Li:08}.
Despite various no-free-lunch theorems \cite{Wolpert:97}, universal
learning is {\em possible} by assuming that the data possess {\em
some} effective structure, but without specifying any further, {\em
which} structure.
Learning algorithms that are universal (at least to some degree) are
also {\em necessary} for developing autonomous general
intelligent systems, required e.g.\ for exploring other planets, as
opposed to decision {\em support} systems which keep a human in the
loop.
There is also an {\em intrinsic} interest in striving for
generality: Finding new learning algorithms for every particular
(new) problem is possible but cumbersome and prone to disagreement
or contradiction. A sound formal general and ideally complete theory
of learning can unify existing approaches, guide the development
of practical learning algorithms, and last but not least lead to
novel and deep insights.

This encyclopedic article gives a mini-introduction into the theory
of universal learning, founded by Ray Solomonoff in the 1960s
\cite{Solomonoff:64,Solomonoff:78} and significantly developed and
extended by the author and his colleagues in the last decade.
It is based on \cite{Hutter:04uaibook}. It explains the spirit of
universal learning, but necessarily glosses over many technical
subtleties. Precise formulation of the results with proofs and/or
references to original publications can be found in
\cite{Hutter:04uaibook}.

\section{Deterministic Environments}\label{secDE}

Let $t,n\in\SetN$ be natural numbers, $\X^*$ be the set of finite
strings and $\X^\infty$ be the set of infinite sequences over some
alphabet $\X$ of size $|\X|$. For a string $x\in\X^*$
of length $\l(x)=n$ we write $x_1x_2...x_n$ with $x_t\in\X$, and
further abbreviate $x_{t:n}:=x_t x_{t+1}...x_{n-1}x_n$ and
$x_{<n}:=x_1... x_{n-1}$, and $\epstr=x_{<1}$ for the empty string.
Consider a countable class of hypotheses $\M=\{H_1,H_2,...\}$. Each
hypothesis $H\in\M$ (also called model) shall describe an infinite
sequence $x_{1:\infty}^H$, e.g.\ like in IQ test questions
``2,4,6,8,....''.
In online learning, for $t=1,2,3,...$, we predict $x_t$ based on
past observations $\dot x_{<t}$, then nature reveals $\dot x_t$, and
so on, where the dot above $x$ indicates the true observation. We
assume that the true hypothesis is in $\M$, i.e.\ $\dot
x_{1:\infty}= x_{1:\infty}^{H_m}$ for some $m\in\SetN$. Goal is to
(``quickly'') identify the unknown $H_m$ from the observations.

\paranodot{Learning by enumeration}
works as follows: Let $\M_t=\{H\in\M:x_{<t}^H=\dot x_{<t}\}$ be the
set of hypotheses consistent with our observations $\dot x_{<t}$ so
far. The hypothesis in $\M_t$ with smallest index, say $m'_t$, is
selected and used for predicting $x_t$.
Then $\dot x_t$ is observed and all $H\in\M_t$ inconsistent with
$x_t$ are eliminated, i.e.\ they are not included in $\M_{t+1}$.
Every prediction error results in the elimination of at least
$H_{m'_t}$, so after at most $m-1$ errors, the true hypothesis $H_m$
gets selected forever, since it never makes an error
($H_m\in\M_t\,\forall t$). This identification may take arbitrarily
long (in $t$), but the number of errors on the way is bounded by
$m-1$, and the latter is often more important.
As an example for which the bound is attained, consider $H_i$ with
$x_{1:\infty}^{H_i}:=1^{f(i)}0^\infty$ $\forall i$ for any
strictly increasing function $f$, e.g.\ $f(i)=i$.
But we now show that we can do much better than this, at least for
finite $\X$.

\paradot{Majority learning}
Consider (temporarily in this paragraph only) a binary alphabet
$\X=\{0,1\}$ and a {\em finite} deterministic hypothesis class
$\M=\{H_1,H_2,...,H_N\}$. $H_m$ and $\M_t$ are as before, but now we
take a majority vote among the hypotheses in $\M_t$ as our
prediction of $x_t$. If the prediction turns out to be wrong, then
at least half (the majority) of the hypotheses get eliminated from
$\M_t$. Hence after at most $\log N$ errors, there is only a single
hypothesis, namely $H_m$, left over. So this majority predictor
makes at most $\log N$ errors.
As an example where this bound is essentially attained, consider
$m=N=2^n-1$ and let $x_{1:\infty}^{H_i}$ be the digits after the comma of
the binary expansion of $(i-1)/2^n$ for $i=1,...,N$.

\paradot{Weighted majority for countable classes}
Majority learning can be adapted to denumerable classes $\M$ and
general finite alphabet $\X$ as follows: Each hypothesis $H_i$ is
assigned a weight $w_i>0$ with $\sum_i w_i\leq 1$. Let
$W:=\sum_{i:H_i\in\M_t}w_i$ be the total weight of the hypotheses in
$\M_t$. Let $\M_t^a:=\{H_i\in\M_t: x_t^{H_i}=a\}$ be the consistent
hypotheses predicting $x_t=a$, and $W_a$ their weight, and take the
weighted majority prediction $x_t=\arg\max_a W_a$. Similarly as
above, a prediction error decreases $W$ by a factor of $1-1/|\X|$,
since $\max_a W_a\geq W/|\X|$. Since $w_m\leq W\leq 1$, this
algorithm can at most make $\log_{1-1/|\X|} w_m=O(\log w_m^{-1})$
prediction errors. If we choose for instance $w_i=(i+1)^{-2}$, the
number of errors is $O(\log m)$, which is an exponential improvement
over the Gold-style learning by enumeration above.

\section{Algorithmic Probability}\label{secAP}

Algorithmic probability has been founded by Solomonoff
\cite{Solomonoff:64}.
The so-called universal probability or a-priori probability is the
key quantity for universal learning. Its philosophical and technical
roots
are \u{Ockham's razor} (choose the simplest model consistent with the data), %
Epicurus' principle of multiple explanations (keep all explanations consistent with the data), %
(Universal) Turing machines (to compute, quantify and assign codes to all quantities of interest), and %
Kolmogorov complexity (to define what simplicity/complexity means).
This section considers deterministic computable sequences, and the
next section the general setup of computable probability
distributions.

\paradot{(Universal) monotone Turing machines}
Since we consider infinite computable sequences, we need devices
that convert input data streams to output data streams. For this we
define the following variants of a classical deterministic Turing
Machine:
A monotone Turing machine $T$ is defined as a Turing machine
with one unidirectional input tape, one unidirectional output tape,
and some bidirectional work tapes. The input tape is binary (no
blank) and read only, the output tape is over finite alphabet $\X$
(no blank) and write only, unidirectional tapes are those where the
head can only move from left to right, work tapes are initially
filled with zeros and the output tape with some fixed element from
$\X$.
We {\em say} that {\em monotone Turing machine} $T$ outputs/computes
a string starting with $x$ on input $p$, and write $T(p)=x*$ if $p$
is to the left of the input head when the last bit of $x$ is output
($T$ reads all of $p$ but no more). $T$ may continue operation and
need not halt. For a given $x$, the set of such $p$ forms a prefix
code. Such codes are called {\em minimal} programs. Similarly we
write $T(p)=\omega$ if $p$ outputs the infinite sequence $\omega$.
%
A {\em prefix code} $\cal P$ is a set of binary strings such that no
element is proper prefix of another. It satisfies {\em Kraft's inequality}
$\sum_{p\in\cal P} 2^{-\l(p)}\leq 1$.

The table of rules of a Turing machine $T$ can be prefix encoded in
a canonical way as a binary string, denoted by $\langle T\rangle$.
Hence, the set of Turing machines $\{T_1,T_2,...\}$ can be
effectively enumerated. There are so-called universal Turing
machines that can ``simulate'' all other Turing machines. We define
a particular one which simulates monotone Turing machine $T(q)$ if
fed with input $\langle T\rangle q$, i.e.\ $U(\langle T \rangle
q)=T(q)$ $\forall T,q$. Note that for $p$ not of the form $\langle
T\rangle q$, $U(p)$ does not output anything.
We call this particular $U$ the {\em reference universal Turing
machine}.

\paradot{Universal weighted majority learning}
$T_1(\epstr), T_2(\epstr), ...$ constitutes an effective
enumeration of all finite and infinite computable sequences, hence
also monotone $U(p)$ for $p\in\{0,1\}^*$. As argued below, the class of
computable infinite sequences, is conceptually very interesting.
The halting problem implies that there is no recursive enumeration of all
partial-recursive functions with infinite domain; hence
we cannot remove the finite sequences algorithmically.
It is very fortunate that we don't have to.
Hypothesis $H_p$ is identified with the sequence $U(p)$,
which may be finite, infinite, or possibly even empty.
The class of considered hypotheses is $\M:=\{H_p:p\in\{0,1\}^*\}$.

The weighted majority algorithm also needs weights $w_p$ for each
$H_p$. Ockham's razor combined with Epicurus' principle demand
to assign a high (low) prior weight to a simple (complex) hypothesis.
If complexity is identified with program length, then $w_p$ should be
a decreasing function of $\l(p)$. It turns out that $w_p=2^{-\l(p)}$
is the ``right'' choice, since minimal $p$ form a prefix code and
therefore $\sum_p w_p\leq 1$ as required.

Using $H_p$ for prediction can now fail in two ways. $H_p$ may make
a wrong prediction or no prediction at all for $x_t$. The true
hypothesis $H_m$ is still assumed to produce an infinite sequence.
The weighted majority algorithm in this setting makes at most
$O(\log w_p^{-1})=O(\l(p))$ errors. It is also plausible
that learning $\l(p)$ bits requires $O(\l(p))$ ``trials''.

\paradot{Universal mixture prediction}
Solomonoff \cite{Solomonoff:78} defined the following universal a-priori probability
\beq\label{Mdef}
  M(x) \;:=\; \sum_{p:U(p)=x*} 2^{-\l(p)}
\eeq
That is, $M(x)=W$ is the total weight of the computable
deterministic hypotheses consistent with $x$ for the universal
weight choice $w_p=2^{-\l(p)}$. The universal weighted majority
algorithm predicted $\arg\max_a M(\dot x_{<t}a)$. Instead, one could
also make a probability prediction $M(a|\dot x_{<t}):=M(\dot
x_{<t}a)/M(\dot x_{<t})$, which is the relative weight of hypotheses
in $\M_t$ predicting $a$. The higher the probability $M(\dot
x_t|\dot x_{<t})$ assigned to the true next observation $\dot x_t$,
the better. Consider the absolute prediction error $|1-M(\dot
x_t|\dot x_{<t})|$ and the logarithmic error $-\log M(\dot x_t|\dot
x_{<t})$.
The cumulative logarithmic error is bounded by $\sum_{t=1}^n
-\log M(\dot x_t|\dot x_{<t}) = -\log M(\dot x_{1:n}) \leq \l(p)$
for any program $p$ that prints $\dot x*$. For instance $p$ could be
chosen as the shortest one printing $\dot x_{1:\infty}$, which has length
$\Km(\dot x_{1:\infty}) := \min\{\l(p):U(p)=\dot x_{1:\infty}\}$.
Using $1-z\leq -\log z$ and letting $n\to\infty$ we get
\beqn
  \sum_{t=1}^\infty |1-M(\dot x_t|\dot x_{<t})|
  \;\leq\; \sum_{t=1}^\infty -\log M(\dot x_t|\dot x_{<t})
  \;\leq\; \Km(\dot x_{1:\infty})
\eeqn
Hence again, the cumulative absolute and logarithmic errors are
bounded by the number of bits required to describe the true
environment.

\section{Universal Bayes}\label{secUB}

The exposition so far has dealt with deterministic environments
only. Data sequences produced by real-world processes are rarely as
clean as IQ test sequences. They are often noisy. This section deals
with stochastic sequences sampled from computable probability
distributions.
%
The developed theory can be regarded as an instantiation of Bayesian
learning. Bayes' theorem allows to update beliefs in face of new
information but is mute about how to choose the prior and the model
class to begin with. Subjective choices based on prior knowledge are
informal, and traditional `objective' choices like Jeffrey's prior
are not universal.
%
Machine learning, the computer science branch of statistics,
develops (fully) automatic inference and decision algorithms for
very large problems. Naturally, machine learning has (re)discovered
and exploited different principles (Ockham's and Epicurus') for
choosing priors, appropriate for this situation.
This leads to an alternative representation of universal probability
as a mixture over all lower semi-computable semimeasures with Kolmogorov
complexity based prior as described below.

\paradot{Bayes}
Sequences $\o=\o_{1:\infty}\in\X^\infty$ are now assumed to be sampled
from the ``true'' probability measure $\mu$, i.e.\
$\mu(x_{1:n}):=\P[\o_{1:n}=x_{1:n}|\mu]$ is the $\mu$-probability
that $\o$ starts with $x_{1:n}$. Expectations w.r.t.\ $\mu$ are denoted
by $\E$. In particular for a function $f:\X^n\to\SetR$, we have
$\E[f]=\E[f(\o_{1:n})]=\sum_{x_{1:n}}\mu(x_{1:n})f(x_{1:n})$.
Note that in Bayesian learning, measures, environments, and models
are the same objects; let
$\M=\{\nu_1,\nu_2,...\}\equiv\{H_{\nu_1},H_{\nu_2},...\}$ denote a
countable class of these measures$\equiv$hypotheses. Assume that
$\mu$ is unknown but known to be a member of ${\cal M}$, and
$w_\nu:=\P[H_\nu]$ is the given prior belief in $H_{\nu}$. Then the
Bayes mixture
\beqn
  \xi(x_{1:n}) \;:=\; \P[\o_{1:n}=x_{1:n}]
  \;=\; \sum_{\nu\in\M}\P[\o_{1:n}=x_{1:n}|H_\nu]\P[H_\nu]
  \;\equiv\; \sum_{\nu\in\M}\nu(x_{1:n})w_\nu
\eeqn
must be our a-priori belief in $x_{1:n}$, and
$\P[H_\nu|\o_{1:n}=x_{1:n}]=w_\nu\nu(x_{1:n})/\xi(x_{1:n})$
be our posterior belief in $\nu$ by Bayes' rule.

\paradot{Universal choice of $\cal M$}
Next we need to find a universal class of environments $\M_U$.
Roughly speaking, Bayes works if $\M$ contains the true environment
$\mu$. The larger $\M$ the less restrictive is this assumption. The
class of all computable distributions, although only countable, is
pretty large from a practical point of view, since it includes for
instance all of today's valid physics theories. (Finding a
non-computable physical system would indeed overturn the
generally accepted Church-Turing thesis.) It is the largest class,
relevant from a computational point of view. Solomonoff
\cite[Eq.(13)]{Solomonoff:64} defined and studied the mixture over
this class.

One problem is that this class is not (effectively=recursively)
enumerable, since the class of computable functions is not
enumerable due to the halting problem, nor is it decidable whether a
function is a measure. Hence $\xi$ is completely incomputable.
Leonid Levin \cite{Zvonkin:70} had the idea to ``slightly'' extend
the class and include also lower semi-computable semimeasures.

A function $\nu:\X^*\to[0,1]$ is a called a semimeasure iff
$\nu(x)\geq\sum_{a\in\X}\nu(xa)\,\forall x\in\X^*$. It is a proper
probability measure iff equality holds and $\nu(\epstr)=1$. $\nu(x)$
still denotes the $\nu$-probability that a sequence starts with
string $x$. A function is called lower semi-computable, if it can be
approximated from below. Similarly to that fact that the class of
partial recursive functions is recursively enumerable, one can show
that the class $\M_U=\{\nu_1,\nu_2,...\}$ of lower semi-computable
semimeasures is recursively enumerable.
In some sense $\M_U$ is the largest class of environments for which
$\xi$ is in some sense computable, but even larger classes are
possible \cite{Schmidhuber:02gtm}.

\paradot{Kolmogorov complexity}
Before we can turn to the prior $w_\nu$, we need to quantify
complexity/simplicity. Intuitively, a string is simple if it
can be described in a few words, like ``the string of one
million ones'', and is complex if there is no such short
description, like for a random object whose shortest
description is specifying it bit by bit. We are interested in
effective descriptions, and hence restrict decoders to be
Turing machines. One can define the {\em prefix Kolmogorov
complexity} of string $x$ as the length $\l$ of the shortest
{\em halting} program $p$ for which $U$ outputs $x$:
\beqn
  K(x) \;:=\; \min_p\{\l(p): U(p)=x\mbox{ halts}\}
\eeqn
Simple strings like 000...0 can be generated by short programs,
and, hence have low Kolmogorov complexity, but irregular (e.g.\
random) strings are their own shortest description, and hence have
high Kolmogorov complexity.
For non-string objects $o$ (like numbers and functions) one defines
$K(o):=K(\langle o\rangle)$, where $\langle o\rangle\in\X^*$ is
some standard code for $o$. In particular, $K(\nu_i)=K(i)$.

To be brief, $K$ is an excellent universal complexity
measure, suitable for quantifying Ockham's razor.

\paradot{The universal prior}
We can now quantify a prior biased towards simple models. First, we
quantify the complexity of an environment $\nu$ or hypothesis
$H_{\nu}$ by its Kolmogorov complexity $K(\nu)$. The universal prior
should be a decreasing function in the model's complexity, and of
course sum to (less than) one. Since $\sum_x 2^{-K(x)}\leq 1$ by the
prefix property and Kraft's inequality, this suggests the choice
\beq\label{uprior}
  w_\nu \;=\; w^U_\nu \;:=\; 2^{-K(\nu)}
\eeq
Since $\log i\leq K(\nu_i)\leq\log i+2\log\log i$ for ``most'' $i$,
most $\nu_i$ have prior approximately reciprocal to their index $i$
as also advocated by Jeffreys and Rissanen.

\paradot{Representations}
Combining the universal class $\M_U$ with the universal prior
\req{uprior}, we arrive at the universal mixture
\beq\label{xiUdef}
  \xi_U(x) \;:=\; \sum_{\nu\in\M_U} 2^{-K(\nu)} \nu(x)
\eeq
which has remarkable properties. First, it is itself a lower
semi-computable semimeasure, that is $\xi_U\in\M_U$, which is
very convenient. Note that for most classes, $\xi\not\in\M$.

Second, $\xi_U$ coincides with $M$ within an irrelevant
multiplicative constant, and $M\in\M_U$. This means that the mixture
over deterministic computable sequences is as rich as the mixture
over the much larger class of semi-computable semimeasures. The
intuitive reason is that the probabilistic semimeasures are in the
convex hull of the deterministic ones, and therefore need not be
taken extra into account in the mixture.

There is another, possibly the simplest, representation: One can
show that $M(x)$ is equal to the probability that $U$ outputs a
string starting with $x$ when provided with uniform random noise on
the program tape.
Note that a uniform distribution is also used in many no-free-lunch theorems
to prove the impossibility of universal learners, but in our case
the uniform distribution is piped through a universal Turing
machine, which defeats these negative implications as we will see
in the next section.

\section{Applications}\label{secAppl}

In the stochastic case, identification of the true hypothesis is
problematic. The posterior $\P[H|x]$ may not concentrate around the
true hypothesis $H_\mu$ if there are other hypotheses $H_\nu$ that
are not asymptotically distinguishable from $H_\mu$. But even if
model identification ({\em induction} in the narrow sense) fails,
{\em predictions}, {\em decisions}, and {\em actions} can be good,
and indeed, for universal learning this is generally the case.

\paradot{Universal sequence prediction}
Given a sequence $x_1x_2...x_{t-1}$,
we want to predict its likely continuation $x_t$. We assume that
the strings which have to be continued are drawn from a computable  ``true''
probability distribution $\mu$.
The maximal prior information a prediction algorithm can possess
is the exact knowledge of $\mu$, but often the true distribution
is unknown. Instead, prediction is based on a guess $\rho$ of
$\mu$. Let $\rho(a|x):=\rho(xa)/\rho(x)$ be the
``predictive'' $\rho$-probability that the next symbol is $a\in\X$,
given sequence $x\in\X^*$. Since $\mu\in\M_U$ it is natural to
use $\xi_U$ or $M$ for prediction.

Solomonoff's \cite{Solomonoff:78,Hutter:04uaibook} celebrated result
indeed shows that $M$ converges to $\mu$. For general alphabet it reads
\beq\label{SolBnd}
  \sum_{t=1}^\infty\E\Big[\sum_{a\in\X}\big(M(a|\o_{<t})-\mu(a|\o_{<t})\big)^2\Big]
  \;\leq\; K(\mu)\ln 2 +O(1)
\eeq
Analogous bounds hold for $\xi_U$ and for other than the Euclidian distance,
e.g.\ the Hellinger and the absolute distance and the relative entropy.

For a sequence $a_1, a_2, ...$ of random variables,
$\sum_{t=1}^\infty\E[a_t^2]\leq c<\infty$ implies $a_t\to 0$ for $t\to\infty$
with $\mu$-probability 1 (w.p.1). Convergence is rapid in the
sense that the probability that $a_t^2$ exceeds $\eps>0$ at more
than $c/\eps\delta$ times, is bounded by $\delta$.
This might loosely be called the number of errors.
%
Hence Solomonoff's  bounds implies
\beqn\label{eqconv2}
  M(x_t|\o_{<t})-\mu(x_t|\o_{<t}) \;\longrightarrow\; 0
  \qmbox{for any $x_t$ rapid w.p.1 for $t\to\infty$}
\eeqn
The number of times, $M$ deviates from $\mu$ by more than $\eps>0$
is bounded by $O(K(\mu))$, i.e.\ is proportional to the complexity
of the environment, which is again reasonable. A counting argument
shows that $O(K(\mu))$ errors for most $\mu$ are unavoidable.
No other choice for $w_\nu$ would lead to significantly better
bounds. Again, in general it is not possible to determine {\em when}
these ``errors'' occur.
%
Multi-step lookahead convergence
$M(x_{t:n_t}|\o_{<t})-\mu(x_{t:n_t}|\o_{<t})\to 0$ even for
unbounded lookahead $n_t-t\geq 0$, relevant for delayed sequence
prediction and in reactive environments, can also be shown.

In summary, $M$ is an excellent sequence predictor under the only
assumption that the observed sequence is drawn from some (unknown)
computable probability distribution. No ergodicity, stationarity, or
identifiability or other assumption is required.

\paradot{Universal sequential decisions}
Predictions usually form the basis for decisions and actions, which
result in some profit or loss. Let $\ell_{x_t y_t}\in[0,1]$ be the
received loss for decision $y_t\in\cal Y$ when $x_t\in\cal X$ turns
out to be the true $t^{th}$ symbol of the sequence.
The $\rho$-optimal strategy
\beqn
  y_t^{\smash{\Lambda_\rho}}(\o_{<t}) \;:=\; \arg\min_{y_t}\sum_{x_t}\rho(x_t|\o_{<t})\ell_{x_t y_t}
\eeqn
minimizes the $\rho$-expected loss.
For instance, if we can decide among $\Y = \{${\it sunglasses}, {\it
umbrella}$\}$ and it turns out to be ${\cal X}=\{\mbox{\it
sun},\mbox{\it rain}\}$, and our personal loss matrix is
$\ell=\left({0.0\;\,0.1\atop 1.0\;\,0.3}\right)$, then
$\Lambda_\rho$ takes $y_t^{\smash{\Lambda_\rho}}=${\it sunglasses}
if $\rho(\mbox{\it rain}|\o_{<t})<\frs18$ and an {\it umbrella}
otherwise.
For ${\cal X}={\cal Y}$ and 0-1 loss $\ell_{xy}=0$ for $x=y$ and 1 else,
$\Lambda_\rho$ predicts the most likely symbol
$y_t^{\smash{\Lambda_\rho}}=\arg\max_a\rho(a|\o_{<t})$ as in Section
\ref{secDE}.

The cumulative $\mu$(=true)-expected loss of $\Lambda_\rho$ for the first $n$
symbols is
\beqn
  \Loss_n^{\smash{\Lambda_\rho}} \;:=\; \sum_{t=1}^n\E[\ell_{\o_t y_t^{\smash{\Lambda_\rho}}(\o_{<t})}]
  \;\equiv\; \sum_{t=1}^n \sum_{x_{1:t}}\mu(x_{1:t})\ell_{x_t y_t^{\smash{\Lambda_\rho}}(x_{<t})}
\eeqn
If $\mu$ is known, $\Lambda_\mu$ obviously results in the best
decisions in the sense of achieving minimal expected loss among all
strategies. For the predictor $\Lambda_M$ based on $M$ (and
similarly $\xi_U$), one can show
\beq\label{lbnd}
  \sqrt{\Loss_n^{\smash{\Lambda_M}}} - \sqrt{\Loss_n^{\smash{\Lambda_\mu}}}
  \;\leq\; \sqrt{2 K(\mu)\ln 2+O(1)}
\eeq
This implies that
$\Loss_n^{\smash{\Lambda_M}}/\Loss_n^{\smash{\Lambda_\mu}}\to 1$ for
$\Loss_n^{\smash{\Lambda_\mu}}\to\infty$, or if
$\Loss_\infty^{\smash{\Lambda_\mu}}$ is finite, then also
$\Loss_\infty^{\smash{\Lambda_M}}<\infty$. This shows that $M$ (via
$\Lambda_M$) performs also excellent from a decision-theoretic
perspective, i.e.\ suffers loss only slightly larger than the
optimal $\Lambda_\mu$ strategy.

One can also show that $\Lambda_M$ is Pareto-optimal
(admissible) in the sense that every other predictor with smaller
loss than $\Lambda_M$ in some environment $\nu\in\M_U$ must
be worse in another environment.

\paradot{Universal classification and regression}
The goal of classification and regression is to infer the functional
relationship $f:\Y\to\X$ from data $\{(y_1,x_1),...,(y_n,x_n)\}$. In
a predictive online setting one wants to ``directly'' infer $x_t$
from $y_t$ given $(y_{<t},x_{<t})$ for $t=1,2,3,...$. The universal
induction framework has to be extended by regarding $y_{1:\infty}$
as independent side-information presented in form of an oracle or
extra tape information or extra parameter. The construction has to
ensure that $x_{1:n}$ only depends on $y_{1:n}$ but is (functionally
or statistically) independent of $y_{n+1:\infty}$.

First, we augment a monotone Turing machine with an extra input tape
containing $y_{1:\infty}$. The Turing machine is called
chronological if it does not read beyond $y_{1:n}$ before $x_{1:n}$
has been written.
Second, semimeasures $\rho=\mu,\nu,M,\xi_U$
are extended to $\rho(x_{1:n}|y_{1:\infty})$, i.e.\ one
semimeasure $\rho(\cdot|y_{1:\infty})$ for each $y_{1:\infty}$ (no
distribution over $y$ is assumed).
Any such semimeasure must be chronological in the sense that
$\rho(x_{1:n}|y_{1:\infty})$ is independent of $y_t$ for $t>n$, hence
we can write $\rho(x_{1:n}|y_{1:n})$.
In classification and regression, $\rho$ is typically
(conditionally) i.i.d., i.e.\
$\rho(x_{1:n}|y_{1:n})=\prod_{t=1}^n\rho(x_t|y_t)$, which is
chronological, but note that the Bayes mixture $\xi$ is {\em not} i.i.d.
One can show that the class of lower semi-computable chronological
semimeasures $\M_U^|=\{\nu_1(\cdot|\cdot),\nu_2(\cdot|\cdot),...\}$
is effectively enumerable.

The generalized universal a-priori semimeasure also has two equivalent definitions:
\beq\label{defMg}
  M(x_{1:n}|y_{1:n}) \;:=\; \sum_{p:U(p,y_{1:n})=x_{1:n}\nq\nq} 2^{-\l(p)}
  \;=\; \sum_{\nu\in\M} 2^{-K(\nu)}\nu(x_{1:n}|y_{1:n})
\eeq
which is again in $\M_U^|$. In case of $|\Y|=1$, this reduces to
\req{Mdef} and \req{xiUdef}. The bounds \req{SolBnd} and \req{lbnd}
and others continue to hold, now for all individual $y$'s,
i.e.\ $M$ predicts asymptotically $x_t$ from $y_t$ and
$(y_{<t},x_{<t})$ for {\em any} $y$, provided $x$ is sampled from a
computable probability measure $\mu(\cdot|y_{1:\infty})$.
Convergence is rapid if $\mu$ is not too complex.

\paradot{Universal reinforcement learning}
The generalized universal a-priori semimeasure \req{defMg} can be
used to construct a universal reinforcement learning agent, called
AIXI. In reinforcement learning, an {\em agent} interacts with an
{\em environment} in cycles $t=1,2,...,n$. In cycle $t$, the agent
chooses an {\em action} $y_t$ (e.g.\ a limb movement) based on past
{\em perceptions} $x_{<t}$ and past actions $y_{<t}$. Thereafter,
the agent perceives $x_t\equiv o_t r_t$, which consists of a (regular)
{\em observation} $o_t$ (e.g.\ a camera image) and a real-valued
{\em reward} $r_t$. The reward may be scarce, e.g.\ just +1
(-1) for winning (losing) a chess game, and 0 at all other times.
Then the next cycle $t+1$ starts. The goal of the agent is to
maximize its expected reward over its lifetime $n$. Probabilistic
planning deals with the situation in which the environmental
probability distribution $\mu(x_{1:n}|y_{1:n})$ is known.
Reinforcement learning deals with the case of unknown $\mu$. In
universal reinforcement learning, the unknown $\mu$ is replaced by
$M$ similarly to the prediction, decision, and classification cases
above. The universally optimal action in cycle $t$ is
\cite{Hutter:04uaibook}
\beq\label{defAIXI}
  y_t \;:=\; \arg\max_{y_t}\sum_{x_t}
   ... \max_{y_n}\sum_{x_n}\, [r_t+...+r_n] M(x_{1:n}|y_{1:n})
\eeq
The expectations ($\Sigma$) and maximizations ($\max$) over future
$x$ and $y$ are interleaved in chronological order to form an
expectimax tree similarly to minimax decision trees in extensive
zero-sum games like chess. Optimality and universality results
similar to the prediction case exist.

\paradot{Approximations and practical applications}
Since $K$ and $M$ are only semi-computable, they have to be
approximated in practice. For instance, $-\log M(x) =
K(x) + O(\log\l(x))$,
and $K(x)$ can and has been approximated by off-the-shelf
compressors like Lempel-Ziv and successfully applied to a plethora
of clustering problems \cite{Cilibrasi:05}. The approximations
upper-bound $K(x)$ and e.g.\ for Lempel-Ziv converge to $K(x)$ if
$x$ is sampled from a context tree source.
The \u{Minimum Description Length principle} \cite{Gruenwald:07book}
also attempts to approximate $K(x)$ for stochastic
$x$.
The Context Tree Weighting algorithm considers a relatively
large subclass of $\M_U$ that can be summed over efficiently.
This can and has been combined with Monte-Carlo sampling
to efficiently approximate AIXI \req{defAIXI} \cite{Hutter:10aixictw}.
The time-bounded versions of $K$ and $M$, namely Levin complexity
$Kt$ and the speed prior $S$ have also been applied to various
learning tasks \cite{Gaglio:07}.

\paradot{Other applications}
Continuously parameterized model classes are very common in
statistics. Bayesian's usually assume a prior {\em density} over
some parameter $\t\in\SetR^d$, which works fine for many problems,
but has its problems. Even for continuous classes $\M$, one can
assign a (proper) universal prior (not density)
$w_\t^U:=2^{-K(\t)}>0$ for computable $\t$ (and $\nu_\t$), and 0 for
uncomputable ones. This effectively reduces $\M$ to a discrete class
$\{\nu_\t\in\M:w_\t^U>0\}\subseteq\M_U$ which is typically dense in
$\M$.
There are various fundamental philosophical and statistical problems
and paradoxes around (Bayesian) induction, which nicely disappear in
the universal framework. For instance, universal induction has no
zero and no improper p(oste)rior problem, i.e.\ can confirm
universally quantified hypotheses, is reparametrization and
representation invariant, and avoids the old-evidence and updating
problem, in contrast to most classical continuous prior densities.
It even performs well in incomputable environments, actually better
than latter \cite{Hutter:07uspx}.

\section{Discussion and Future Directions}\label{secDisc}

Universal learning is designed to work for a wide range of problems
without any a-priori knowledge.
In practice we often have extra information about the problem at
hand, which could and should be used to guide the forecasting.
One can incorporate it by explicating all our prior
knowledge $z$, and place it on an extra input tape of our universal
Turing machine $U$, or prefix our observation sequence $x$ by $z$ and
use $M(zx)$ for prediction.

Another concern is the dependence of $K$ and $M$ on $U$. The good
news is that a change of $U$ changes $K(x)$ only within an additive
and $M(x)$ within a multiplicative constant independent of $x$. This
makes the theory practically immune to any ``reasonable'' choice of
$U$ for large data sets $x$, but predictions for short sequences
(shorter than typical compiler lengths) can be arbitrary. One
solution is to take into account our (whole) scientific prior
knowledge $z$ \cite{Hutter:06hprize}, and predicting the now long
string $zx$ leads to good (less sensitive to ``reasonable'' $U$)
predictions. This is a kind of grand transfer learning scheme. It is
unclear whether a more elegant theoretical solution is possible.

Finally, the incomputability of $K$ and $M$ prevents a {\em direct}
implementation of Solomonoff induction. Most fundamental theories
have to be approximated for practical use, sometimes systematically
like polynomial time approximation algorithms or numerical
integration, and sometimes heuristically like in many AI-search
problems or in non-convex optimization problems. Universal machine
learning is similar, except that its core quantities are only
semi-computable. This makes them often hard, but as described in the
previous section, not impossible, to approximate.

In any case, universal induction can serve as a ``gold standard''
which practitioners can aim at.
Solomonoff's theory considers the class of all computable
(stochastic) models, and a universal prior inspired by Ockham and
Epicurus, quantified by Kolmogorov complexity. This lead to a
universal theory of induction, prediction, decisions, and, by
including Bellman, to universal actions in reactive environments.
Future progress on the issues above (incorporating prior knowledge,
getting rid of the compiler constants, and finding better
approximations) will lead to new insights and will continually
increase the number of applications.

\newpage
\def\refname{Recommended Reading}
{\addcontentsline{toc}{section}{\refname}

\begin{small}

\end{small}

\end{document}